\documentclass[final]{cvpr}

\usepackage{times}
\usepackage{epsfig}
\usepackage{graphicx}
\usepackage{amsmath}
\usepackage{amssymb}

\usepackage{subcaption}
\usepackage{placeins}

\usepackage{booktabs}
\usepackage{bbm}

\makeatletter
\@namedef{ver@everyshi.sty}{}
\makeatother
\usepackage{tikz}

\usepackage{tikzsymbols}
\usetikzlibrary{arrows}
\usetikzlibrary{chains}
\usetikzlibrary{patterns}
\usepackage{tikz-qtree}
\usetikzlibrary{shapes.multipart}
\usetikzlibrary{positioning}
\usetikzlibrary{arrows.meta}
\usepackage{pgfplots}

\usetikzlibrary{calc}
\usetikzlibrary{positioning}
\usepgflibrary{fpu}
\usepgfplotslibrary{fillbetween}

\usetikzlibrary{external}
\pgfplotsset{compat=newest}

\usepackage{amsmath}
\usepackage{amssymb}

\usepackage{array}

\DeclareMathOperator*{\argmax}{arg\,max}

\usepackage[pagebackref=true,breaklinks=true,colorlinks,bookmarks=false]{hyperref}

\def\cvprPaperID{40} 
\def\confYear{CVPR 2021}

\begin{document}

\title{I Find Your Lack of Uncertainty in Computer Vision Disturbing}

\author{Matias Valdenegro-Toro\\
German Research Center for Artificial Intelligence\\
Robert-Hooke-Str 1, 28359 Bremen, Germany\\
{\tt\small matias.valdenegro@dfki.de}
}

\maketitle

\begin{abstract}
   Neural networks are used for many real world applications, but often they have problems estimating their own confidence. This is particularly problematic for computer vision applications aimed at making high stakes decisions with humans and their lives.
   In this paper we make a meta-analysis of the literature, showing that most if not all computer vision applications do not use proper epistemic uncertainty quantification, which means that these models ignore their own limitations.
   We describe the consequences of using models without proper uncertainty quantification, and motivate the community to adopt versions of the models they use that have proper calibrated epistemic uncertainty, in order to enable out of distribution detection. We close the paper with a summary of challenges on estimating uncertainty for computer vision applications and recommendations.
\end{abstract}

\section{Introduction}

Neural networks are currently used in many real-world applications, from medical systems to autonomous vehicles \cite{feng2020review}, robots \cite{de2017multimodal} and underwater manipulator systems \cite{hildebrandt2008computer}. Despite these advances, there are many practical issues, for example, fairness and transparency in automated decisions \cite{buolamwini2018gender}, decision and prediction interpretability, and properly estimating epistemic uncertainty to preventing a model from making overconfident predictions \cite{guo2017calibration}. These issues are of increasing importance as machine learning models are used to make decisions in the real world, affecting humans and their lives \cite{feng2020review}.

In this line we consider the uncertainty of model outputs, and how they are often overlooked while the field of computer vision makes progress. The purpose of uncertainty quantification in a machine learning model is to produce good estimates of predictive uncertainty, with the aims to help the user identify which predictions are trustworthy and which ones are unreliable.

\begin{figure}
    \includegraphics[width=\linewidth]{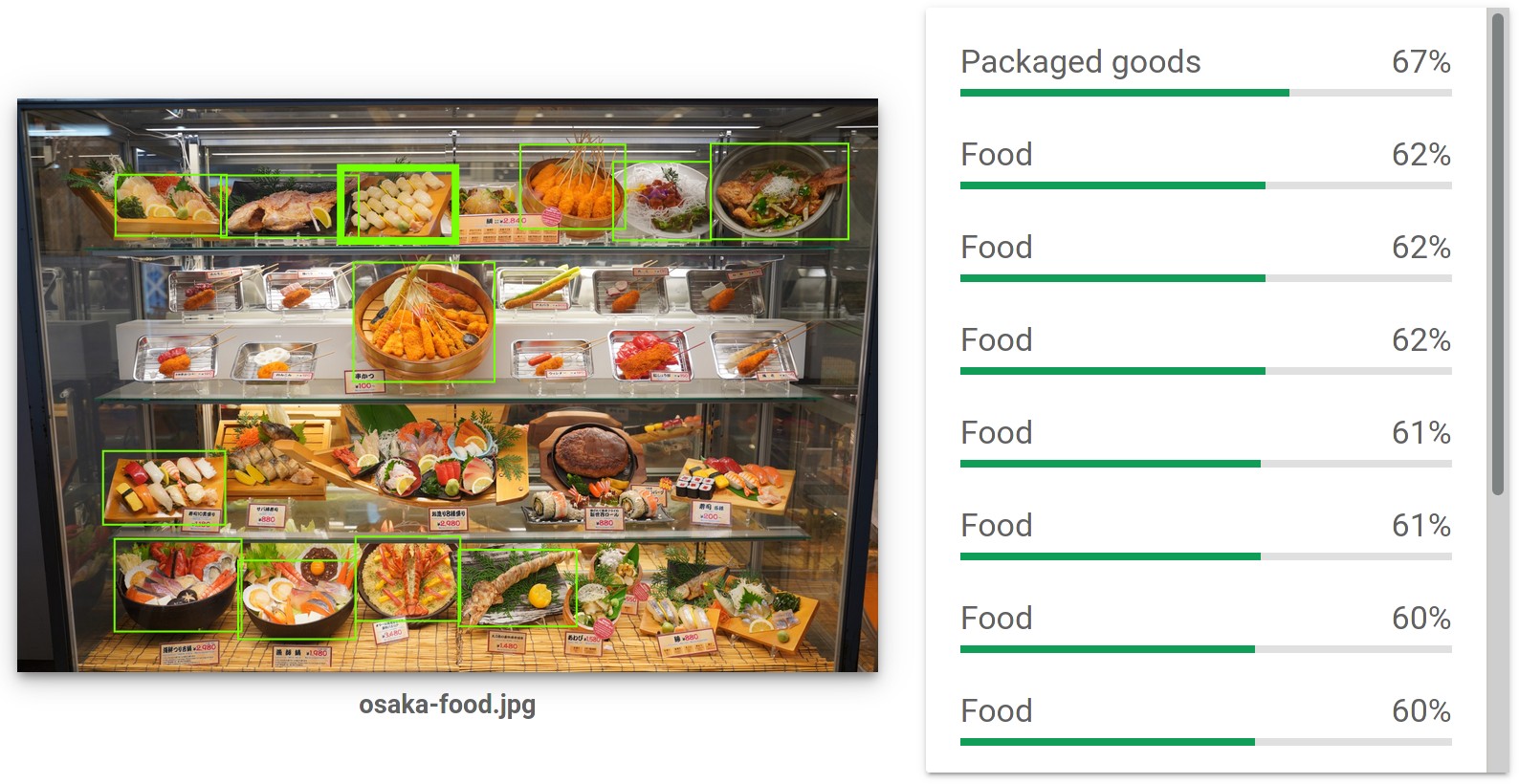}
    \caption{Computer vision models that use machine learning do not often correctly estimate their own uncertainty. In this example, the model is only around 60\% confident on these food samples, while confidence should be closer to 100\%. Note that some objects are not detected. Results obtained from Google Cloud Vision AI.}
    \label{osakaFoodGoogleVisionAI}
\end{figure}

But neural network models often produce overconfident incorrect predictions \cite{guo2017calibration}, where the prediction is incorrect but the model confidence is very high. Overconfidence produce a false sense of security, as the model produces higher confidences than it should, and the user is misled to trust incorrect predictions. Confidences are usually interpreted by a human and should "make sense" \cite{cosmides1996humans}. Underconfidence can also be a problem, as shown in Figure \ref{osakaFoodGoogleVisionAI}.

In this paper, we perform a meta-analysis of computer vision applications that are aimed or used in real-world scenarios, and show that most of them do not consider uncertainty quantification as an integral part of their system design. We believe that model without proper uncertainty are dangerous to humans and have potential ethical and legal concerns. Our work only considers computer vision methods that are backed by machine learning models, which is the standard paradigm in modern computer vision.

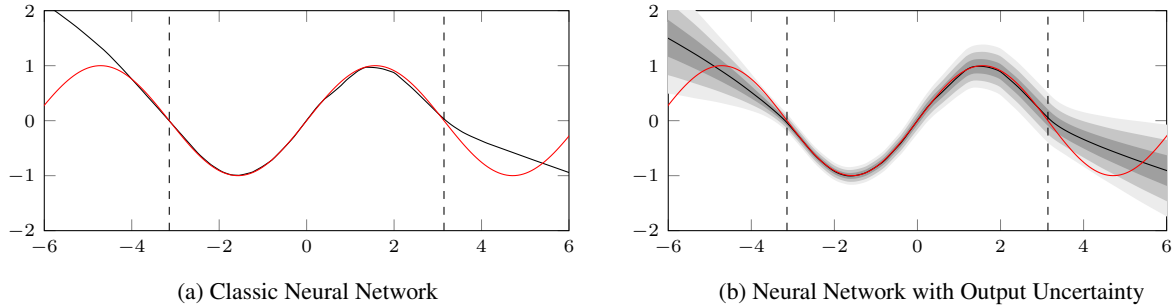
\begin{figure*}[t]
    \centering
    \begin{subfigure}{0.47 \textwidth}
        \begin{tikzpicture}
            \begin{axis}[height = 0.20 \textheight, width = \linewidth, xlabel={}, ylabel={}, xmin = -6.0, xmax = 6.0, grid style=dashed, legend pos = north east, legend style={font=\scriptsize}, tick label style={font=\scriptsize}, ymin =-2.0 , ymax=2.0]
                
                \addplot+[mark = none, black] table[x  = x, y  = pred_mu, col sep = semicolon] {experiments/regression/toy-regression-deepensembles-1.csv};
                \addplot+[mark = none, red] table[x  = x, y  = true_mu, col sep = semicolon] {experiments/regression/toy-regression-deepensembles-1.csv};
                
                \draw[dashed] (3.1415, -2) -- (3.1415, 2);
                \draw[dashed] (-3.1415, -2) -- (-3.1415, 2);
            \end{axis}		
        \end{tikzpicture}
        \caption{Classic Neural Network}
    \end{subfigure}
    \begin{subfigure}{0.47 \textwidth}
        \begin{tikzpicture}
            \begin{axis}[height = 0.20 \textheight, width = \linewidth, xlabel={}, ylabel={}, xmin = -6.0, xmax = 6.0, grid style=dashed, legend pos = north east, legend style={font=\scriptsize}, tick label style={font=\scriptsize}, ymin =-2.0 , ymax=2.0]
                
                \addplot+[mark = none, black] table[x  = x, y  = pred_mu, col sep = semicolon] {experiments/regression/toy-regression-deepensembles-15.csv};
                \addplot+[mark = none, red] table[x  = x, y  = true_mu, col sep = semicolon] {experiments/regression/toy-regression-deepensembles-15.csv};

                \addplot[name path=upper, draw=none] table[x = x, y expr = \thisrow{pred_mu} + \thisrow{pred_sigma}, col sep = semicolon] {experiments/regression/toy-regression-deepensembles-15.csv};
                \addplot[name path=lower, draw=none] table[x = x, y expr = \thisrow{pred_mu} - \thisrow{pred_sigma}, col sep = semicolon] {experiments/regression/toy-regression-deepensembles-15.csv};

                \addplot[name path=upperTwo, draw=none] table[x = x, y expr = \thisrow{pred_mu} + 2 *\thisrow{pred_sigma}, col sep = semicolon] {experiments/regression/toy-regression-deepensembles-15.csv};
                \addplot[name path=lowerTwo, draw=none] table[x = x, y expr = \thisrow{pred_mu} - 2 *\thisrow{pred_sigma}, col sep = semicolon] {experiments/regression/toy-regression-deepensembles-15.csv};
                
                \addplot[name path=upperThree, draw=none] table[x = x, y expr = \thisrow{pred_mu} + 3 *\thisrow{pred_sigma}, col sep = semicolon] {experiments/regression/toy-regression-deepensembles-15.csv};
                \addplot[name path=lowerThree, draw=none] table[x = x, y expr = \thisrow{pred_mu} - 3 *\thisrow{pred_sigma}, col sep = semicolon] {experiments/regression/toy-regression-deepensembles-15.csv};
                
                \addplot [fill=darkgray!10] fill between[of=upperThree and lowerThree]; 
                \addplot [fill=darkgray!30] fill between[of=upperTwo and lowerTwo]; 
                \addplot [fill=darkgray!50] fill between[of=upper and lower];

                \draw[dashed] (3.1415, -2) -- (3.1415, 2);
                \draw[dashed] (-3.1415, -2) -- (-3.1415, 2);       
            \end{axis}		
        \end{tikzpicture}
        \caption{Neural Network with Output Uncertainty}
    \end{subfigure}    
    \caption{Comparison between classic and neural networks with output uncertainty for regression of $f(x) =\sin(x) + \epsilon $ where $\epsilon \sim \mathcal{N}(0, \sigma^2(x))$, $\sigma(x) = 0.15(1 + e^{-x})$, and the training set is $x \in [-\pi, \pi]$. The shaded areas represent one to three $\sigma$ confidence intervals. Dotted lines indicate the limits of the training set. The output of the classic NN does not indicate any anomaly when predicting incorrectly outside of the training set, while the NN with uncertainty indicates extrapolation through the increasing variance output head.}
    \label{toyRegressionComparison}
\end{figure*}

Probabilities produced by a softmax activation are generally uncalibrated \cite{guo2017calibration} and can lead to overconfident incorrect predictions, giving a false sense of security if these confidences are used for further decision making. An appropriate model would produce calibrated probabilities and well behaved output distributions. This applies both to classification and regression problems (and multi-task combinations like object detection \cite{hall2020probabilistic}).

We point that there is a large gap between applications using machine learning and bayesian or uncertainty quantification methods that are able to estimate uncertainty in a principled way. The lesson that we want to strongly leave on the reader is:

\begin{center}
    \emph{All applications using computer vision systems should use \textbf{only} methods producing calibrated epistemic uncertainty}
\end{center}

\FloatBarrier

Even as some computer vision methods have probabilistic variants with good uncertainty properties \cite{hall2020probabilistic} \cite{feng2020review}, these are not generally used in end user applications. There have been large advances in bayesian and uncertainty quantification methods for computer vision \cite{gustafsson2020evaluating}, there are still gaps in its use on real-world applications.

Many application fields would benefit from correct modeling of uncertainty in inputs and outputs, such as systems making medical decisions, robotics, and autonomous driving. What is common to these fields is the involvement and interaction with humans, and the potential risk of acting incorrectly in unexpected situations.

\section{What is Uncertainty?}

Uncertainty is a measure for lack of information, or impossibility to exactly describe a state or future outcomes. In machine learning models and computer vision systems, uncertainty is generally translated as confidence scores for different predictions: class probabilities, standard deviation for continuous outputs, etc. These indicate the level of certainty in the system given some input data.

There are two basic types of uncertainty \cite{der2009aleatory} as defined by its source:

\textbf{Aleatoric Uncertainty}. It is the one inherent to the data and how it is captured, for example measurement uncertainty, stochastic processes, etc. It cannot be reduced by incorporating additional data. \cite{hullermeier2021aleatoric} In computer vision contexts, this kind of uncertainty can be annotation noise (multiple correct classes or bounding box uncertainty), camera noise that translates into uncertain pixels, etc.

\textbf{Epistemic Uncertainty}. This kind is inherent to the model and its processes, it can be reduced by adding more information, such as additional training data, better model specification, or improved methods \cite{hullermeier2021aleatoric}. This kind of uncertainty is most useful for out of distribution detection \cite{hendrycks2017baseline}. In computer vision contexts, some examples are visual similarity between classes, bounding box localization uncertainty, etc.

An example of both kinds of uncertainty is shown in a regression setting in Figure \ref{toyRegressionComparison}. The left plot shows that the classic model works well inside the training set, but it produces incorrect predictions outside of the training set, with no indication of extrapolation. The right plot shows the output of a Deep Ensemble \cite{lakshminarayanan2017simple}, where the variance output of the model increases significantly when the input is outside of the training set range, which indicates that the model is extrapolating. Aleatoric uncertainty can be seen inside the training set, as the noise variance varies with the input ($\sigma(x) = 0.15(1 + e^{-x})$) and the output variance increase is consistent with the noise term.

\subsection{Bayesian Formulation}

The Bayesian neural network (BNN) is the standard formulation for uncertainty quantification in neural networks \cite{mackay1992evidence}, with many methods being approximations of the full BNN through different techniques, such as ensembling \cite{lakshminarayanan2017simple} or variational inference \cite{blundell2015weight}.

In a BNN, the weights of each layer are probability distributions $P(w)$ that are learned in the training process. We do not cover the learning of these distributions. Once learned, predictions can be made using the posterior predictive distribution shown below:

\begin{equation}
    P(y \,|\, x) = \int_w P(y \,|\, x, w) P(w) dw
\end{equation}

Standard machine learning methods usually are not able to provide high quality uncertainty estimation, since they only produce point predictions, while BNNs output probability distributions, which can encode predictive uncertainty directly. A BNN works by propagating the uncertainty in the parameters (weights) through the network, and given an input, the computed output is also a probability distribution that represents predictive uncertainty.

\subsection{Calibration}

An important concept in uncertainty quantification is how to evaluate the quality of the produced confidence scores. We expect that low confidence predictions have a tendency to be incorrect, and high confidence predictions should be correct most of the time. This can be quantified with the concept of reliability diagrams and different metrics that can be computed over them.

A reliability diagram is built by taking the class predictions $c$ and confidence/probabilities $p_i$, where $c = \argmax_i p_i$, $i$ is an index over classes, and per-class confidence is $p = \max_i p_i$, then split the predictions by confidence values $p$ into bins $B_i$, for each bin the accuracy $\text{acc}(B_i) = |B_i|^{-1} \sum_{j \in B_i} \mathbbm{1}[c_j = t_j]$ is computed, then the mean confidence for each bin is computed as $\text{conf}(B_i) = \sum_{j \in B_i} p_j$ and then the values $(\text{conf}(B_i), \text{acc}(B_i))$ are plotted. Note that bins are usually equally spaced, and the number of bins is a hyper-parameter that must be selected carefully.

Regions where $\text{conf}(B_i) < \text{acc}(B_i)$ indicate that the model is underconfident, while regions $\text{conf}(B_i) > \text{acc}(B_i)$ indicate overconfidence. The line $\text{conf}(B_i) = \text{acc}(B_i)$ indicates perfect calibration. Examples and their interpretation are shown in Figure \ref{reliabilityPlots}. Some metrics that can be computed from a reliability plot are:

\begin{description}
    \item[Calibration error \cite{guo2017calibration}]
    $$\text{CE} = \sum_i |\text{acc}(B_i) - \text{conf}(B_i)|$$
    \item[Expected Calibration error \cite{naeini2015obtaining}]
    $$\text{ECE} = \sum_i N^{-1} |B_i|  \left|\text{acc}(B_i) - \text{conf}(B_i)\right|$$
    \item[Maximum Calibration error \cite{naeini2015obtaining}]
    $$\text{MCE} = \max_i \left|\text{acc}(B_i) - \text{conf}(B_i)\right|$$
\end{description}

These metrics are used to numerically evaluate the calibration properties of a model. There are definitions of calibration metrics for regression problems, where accuracy is replaced by $1 - \alpha$ confidence intervals containing the true values \cite{kuleshov2018accurate}. Models can be calibrated after training \cite{platt1999probabilistic}, but generally this does not produce high quality uncertainty \cite{ovadia2019trust}

\newcommand{\drawge}{-- (rel axis cs:1,0) -- (rel axis cs:1,1) -- (rel axis cs:0,1) \closedcycle}
\newcommand{\drawle}{-- (rel axis cs:1,1) -- (rel axis cs:1,0) -- (rel axis cs:0,0) \closedcycle}

\begin{figure}
    \centering
    \begin{tikzpicture}
        \begin{axis}[height = 0.15 \textheight, width = 0.40 \linewidth, xlabel={Confidence}, ylabel={Accuracy}, xmin = 0.0, xmax = 1.0, ymin = 0.0, ymax = 1.0, ymajorgrids=false, xmajorgrids=false, grid style=dashed, legend pos = north east, legend style={font=\scriptsize}, x tick label style={font=\tiny, rotate=90}, y tick label style={font=\tiny}, title={Overconfident Model}]
            
            \addplot+[mark = none, gray, dashed] coordinates { (0, 0) (1, 1)};
            \addplot+[mark = none, blue!30,ybar,fill] coordinates { (0.0,0.0) (0.1, 0.05) (0.2, 0.15) (0.3, 0.2) (0.4, 0.15) (0.5, 0.35) (0.6, 0.48) (0.7, 0.55) (0.8, 0.5) (0.9, 0.8)};
            
        \end{axis}		
    \end{tikzpicture}
    \begin{tikzpicture}
        \begin{axis}[height = 0.15 \textheight, width = 0.40 \linewidth, xlabel={Confidence}, ylabel={Accuracy}, xmin = 0.0, xmax = 1.0, ymin = 0.0, ymax = 1.0, ymajorgrids=false, xmajorgrids=false, grid style=dashed, legend pos = north east, legend style={font=\scriptsize}, x tick label style={font=\tiny, rotate=90}, y tick label style={font=\tiny}, title={Underconfident Model}]                
            
            \addplot+[mark = none, blue!30,ybar,fill] coordinates { (0.0,0.0) (0.1, 0.1) (0.2, 0.3) (0.3, 0.4) (0.4, 0.35) (0.5, 0.55) (0.6, 0.8) (0.7, 0.75) (0.8, 0.8) (0.9, 0.95)};
            \addplot+[mark = none, gray, dashed] coordinates { (0, 0) (1, 1)};
            
        \end{axis}		
    \end{tikzpicture}

    \begin{tikzpicture}
        \begin{axis}[height = 0.18 \textheight, width = 0.54 \linewidth, xlabel={Confidence}, ylabel={Accuracy}, xmin = 0.0, xmax = 1.0, ymin = 0.0, ymax = 1.0, ymajorgrids=false, xmajorgrids=false, grid style=dashed, legend pos = north east, legend style={font=\scriptsize}, x tick label style={font=\scriptsize, rotate=90}, y tick label style={font=\scriptsize},ticks=none]
            
            \addplot+[mark = none, gray, dashed] coordinates { (0, 0) (1, 1)};
            \addplot[draw=none, fill, blue!40, domain=0:1] { x} \drawge;
            \addplot[draw=none, fill, red!40, domain=0:1] { x} \drawle;
            
            \node[] at (axis cs: 0.4,0.9) {\small Underconfidence};
            \node[] at (axis cs: 0.6,0.1) {\small Overconfidence};
            
       \end{axis}		
    \end{tikzpicture}
    \caption{Reliability plots and its interpretation to evaluate underconfidence or overconfidence on a model's predictions.}
    \label{reliabilityPlots}
\end{figure}
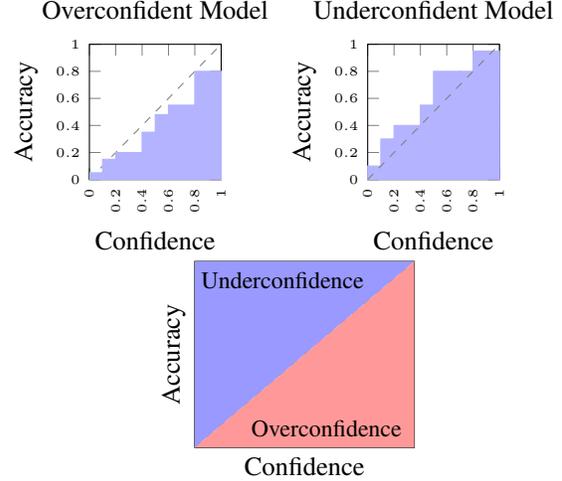

\section{Computer Vision Models in Real-World Applications}

We now make a small survey of some real-world applications of computer vision, which standard models they use, and what kind of uncertainty quantification is performed. We used multiple criteria, including popularity of the computer vision method, societal interests (like medical or agricultural applications), and applications that involve humans and their environment (such as robots). We decided to leave out any research work that presents its main results on simulation or non-realistic data, as we would like to emphasize real-world applications using computer vision methods.

Table \ref{surveyCVTasksRealWorld} presents our survey results. Most applications using computer vision methods do not use any kind of uncertainty quantification. The only exceptions are some methods that produce uncalibrated probabilities, such as image classifiers, or , but generally these are only partial estimates of aleatoric uncertainty.

In Table \ref{uncertaintyModelComparison} we also show that many of the popular computer vision methods have bayesian or probabilistic variants in the literature. We evaluated each method according to if they produce aleatoric, epistemic, or both kinds of uncertainty. For aleatoric uncertainty, we usually look if the method outputs a probability distribution and aleatoric uncertainty is considered in their loss function. Epistemic uncertainty requires bayesian neural network modeling (or approximations like MC-Dropout) using distributions over weights, or ensembles of the same architecture.

We would like to motivate the community to prefer the use of models with appropriate uncertainty quantification.

\begin{table*}[t]
    \centering
    \begin{tabular}{llllp{1.5cm}p{4.0cm}}
        \toprule
        Domain Task & Ref & Method & Uncertainty? & Challenges & Comments\\
        \midrule
        \textbf{Object Detection}		& \\
        Breast Cancer Detection & \cite{mckinney2020international}	& Faster R-CNN & Ensemble & B C E G & Ensembles for performance improvement\\
        Fruit Detection for Robots & \cite{wan2020faster} & Faster R-CNN & None & A D G & \\
        \midrule
        \textbf{Semantic Segmentation}	& \\
        Cell Tracking & \cite{ronneberger2015u}	& U-Net & None & G E & \\
        Robot Navigation & \cite{valada2017adapnet}		& AdapNet FCN & None & A C D G & \\ 
        \midrule
        \textbf{Instance Segmentation}	& \\
        Fruit Detection for Robots & \cite{yu2019fruit} & Mask R-CNN & None & A D G & \\
        Tree Detection for AUVs & \cite{ocer2020tree}	& Mask R-CNN & None & A D G H & \\
        \midrule
        \textbf{Image Classification}\\					  
        COVID19 Detection &							  & AlexNet & None & \\
        from Chest X-Ray & \cite{abbas2021classification} & VGG19 & None &  B C E G H & Full survey at \cite{roberts2021common}\\
        Facial Emotion Recognition & \cite{arriaga2017real} & mini-Xception & None & C B F G & \\
        \midrule
        \textbf{Pose Estimation}		& \\
        Human Pose Estimation & \cite{cao2019openpose} & OpenPose & None & A D E G H& Uncalibrated confidence maps for human parts\\
        Multimodal Driver\\
        Behavior Recognition & \cite{martin2019drive} & OpenPose & None &  E G H & \\
        \midrule
        \textbf{Tracking}				& \\
        Pedestrian Tracking & \cite{goccer2019pedestrian} & Faster R-CNN & None & A E G & \\
        \midrule
        \textbf{Feature Learning}		& \\
        Facial Feature Learning & \cite{schroff2015facenet} & FaceNet & None & E F G & Embedding distances could be an (uncalibrated) uncertainty measure.\\ 
        \midrule
        \textbf{Various}		& \\
        Autonomous Driving & \cite{bojarski2016end}	 & Multiple & None & A-H (All) & \\
        Drone Obstacle Avoidance & \cite{loquercio2018dronet} & ResNet & None & A E D G & Uncalibrated collision probabilities.\\
        Grasp Point Prediction & \cite{pinto2016supersizing} & Multi-Label & None & A B F G & Uncalibrated angle probabilities\\
        & & AlexNet & & \\
        Vision \& Language Robot Navigation & \cite{anderson2020sim} & Custom & None & A E G & \\
        \bottomrule
    \end{tabular}
    \vspace*{0.1em}
    \caption{Selection of computer vision tasks and their use in real-world applications, including information on any uncertainty quantification methods used. The challenges column relates to the future challenges presented in Section 8.}
    \label{surveyCVTasksRealWorld}
\end{table*}

\section{The Sins of Uncertainty in CV}

We believe that there are three basic mistakes of uncertainty quantification (UQ) in computer vision.

\textbf{Not Using UQ in CV}. Models without any kind of explicit uncertainty quantification will produce uncalibrated probabilities \cite{guo2017calibration} and generally only make point predictions, which cannot perform high quality uncertainty quantification. This translates in producing incorrect overconfident predictions and not being able to use prediction confidence to detect misclassified and out of distribution examples \cite{hendrycks2017baseline}.

\textbf{Using Inappropriate UQ Methods}. When using a method for uncertainty quantification, the type of uncertainty that the model can produce should be considered. Many methods can only estimate aleatoric uncertainty, while not being able to estimate epistemic uncertainty. The latter kind is required for tasks such as out of distribution detection, and is generally more interesting for estimating if the model is performing outside of its training distribution \cite{tagasovska2019single}. Aleatoric uncertainty is easier to estimate with an additional output head that estimates variance of the data, with an appropriate loss function, but epistemic uncertainty requires additional modeling such as BNNs with a distribution over the weights \cite{mackay1992evidence}, or multiple models like ensembling \cite{lakshminarayanan2017simple}.

An important issue is to disentangle epistemic from aleatoric uncertainty. Some methods can produce separate estimates (like Deep Ensembles \cite{lakshminarayanan2017simple} for regression, or SDE-Net \cite{kong2020sde}, or frameworks like \cite{kendall2017uncertainties}), while others can provide a single estimate that combines both epistemic and aleatoric uncertainty. This is generally called predictive uncertainty \cite{detlefsen2019reliable}.

\begin{table}
    \centering
    \begin{tabular}{lllll}
        \toprule
        CV 		  		& Probabilistic  			& Ref & AU 	& EU \\
        Method    		& Variant\\
        \midrule
        YOLOv3	  		& GaussianYOLOv3 & \cite{choi2019gaussian}			& Yes	& No\\
        \midrule
        Faster   	 	& Soft-NMS FRCNN & \cite{he2019bounding}				& Yes 	& No \\
        R-CNN & MC Dropout FRCNN & \cite{miller2019benchmarking}	& Yes 	& Yes\\
        & Ensemble FRCNN & \cite{miller2019benchmarking}		& Yes 	& Yes\\
        SSD				& MC Dropout SSD & \cite{miller2019benchmarking}		& Yes 	& Yes\\
        & Ensemble SSD & \cite{miller2019benchmarking}		& Yes 	& Yes\\
        \midrule
        RetinaNet		& BayesOD & \cite{harakeh2020bayesod}					& Yes	& Yes\\
        \midrule
        Mask 	 		& Soft-NMS & \\
        R-CNN			& Mask R-CNN & \cite{he2019bounding}		    & Yes 	& No \\
        \midrule
        SegNet			& Bayesian SegNet & \cite{kendall2015bayesian} & Yes & Yes\\
        \bottomrule
    \end{tabular}
    \vspace*{0.1em}
    \caption{Comparison between classic computer vision methods and their probabilistic/bayesian versions, including information on uncertainty quality: aleatoric (AU) and epistemic (EU).}
    \label{uncertaintyModelComparison}
\end{table}

\textbf{Using UQ but not acting on Uncertainty}. Uncertainty quantification can be a powerful method to improve performance, depending on the method. For example, ensembles \cite{lakshminarayanan2017simple} have good uncertainty quantification properties, but additionally to this they also improve performance on the original task, motivating its use for purposes other than quantifying uncertainty.

A good example of this oversight is \cite{mckinney2020international}, where a three member ensemble of object detectors is used as a way to improve the overall performance of the system, but without considering that disagreement between ensemble members can be used as an uncertainty measure that provides additional information to the user (most likely a physician).

Uncertainty quantification method should not be used only with the purpose of improving on a benchmark. There is much more that can be obtained than just task performance. For example, performing and evaluating out of distribution detection to know the model's limits, or just as additional information to the user.

\section{The Consequences of Overconfidence}

\begin{figure*}
    \centering
    \includegraphics[width=0.46\textwidth]{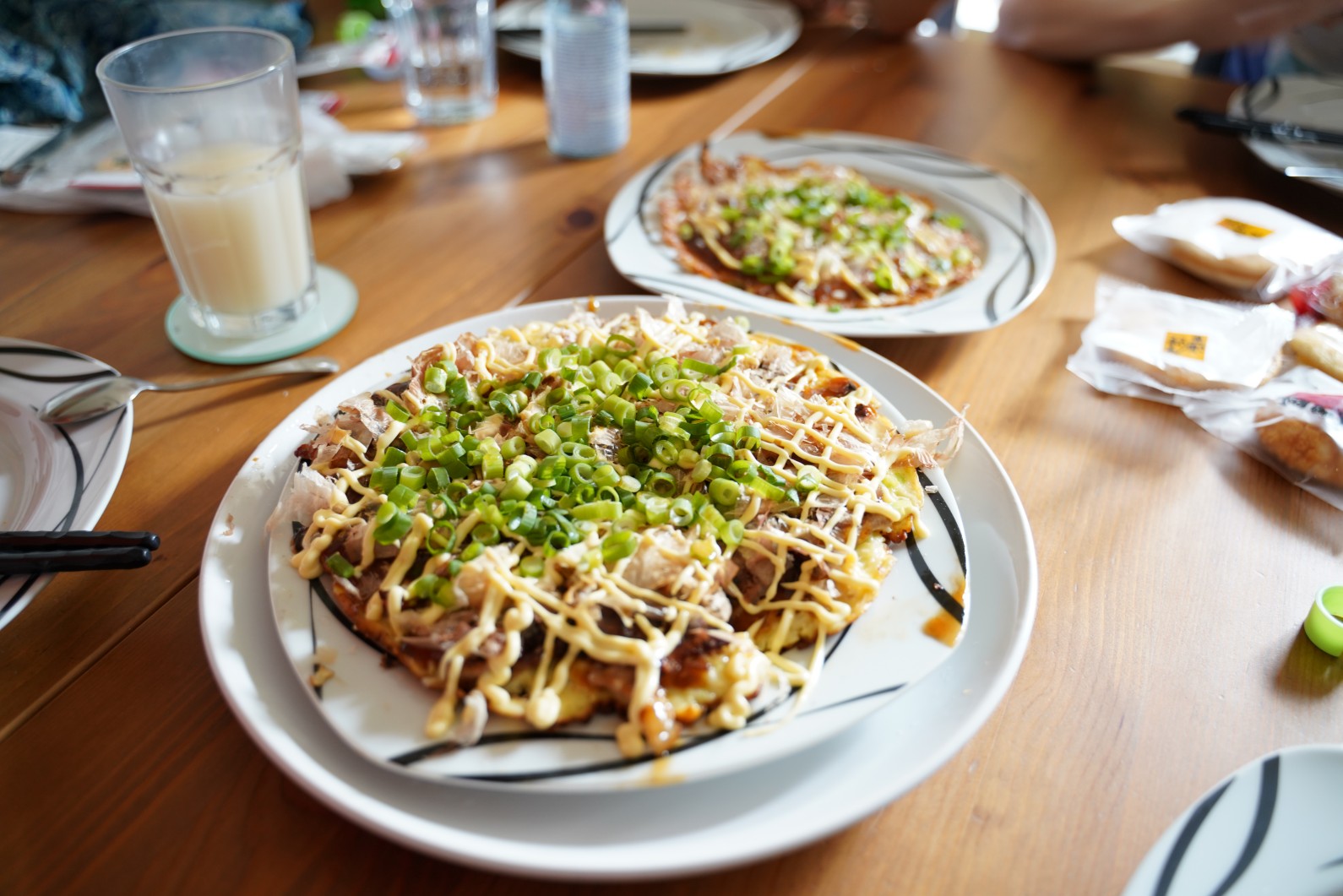}
    \includegraphics[width=0.46\textwidth]{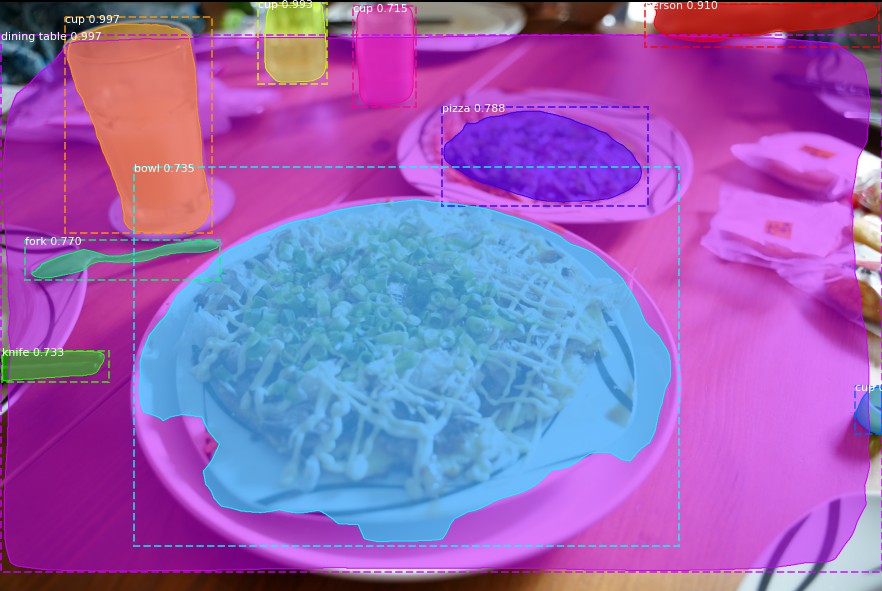}
    \caption{Incorrect detections with Mask R-CNN trained on COCO. Left presents the input image, and right the predictions. The Okonomiyaki is misclassified as a Bowl and a Pizza, and a Spoon is misclassified as a Fork, all with high confidence of $\sim 75\%$. The Bowl-Okonomiyaki has high variability in the predicted mask boundaries, signaling some amount of uncertainty.}
    \label{maskRCNNOkonomiyaki}
\end{figure*}

\begin{figure}
    \includegraphics[width=\linewidth]{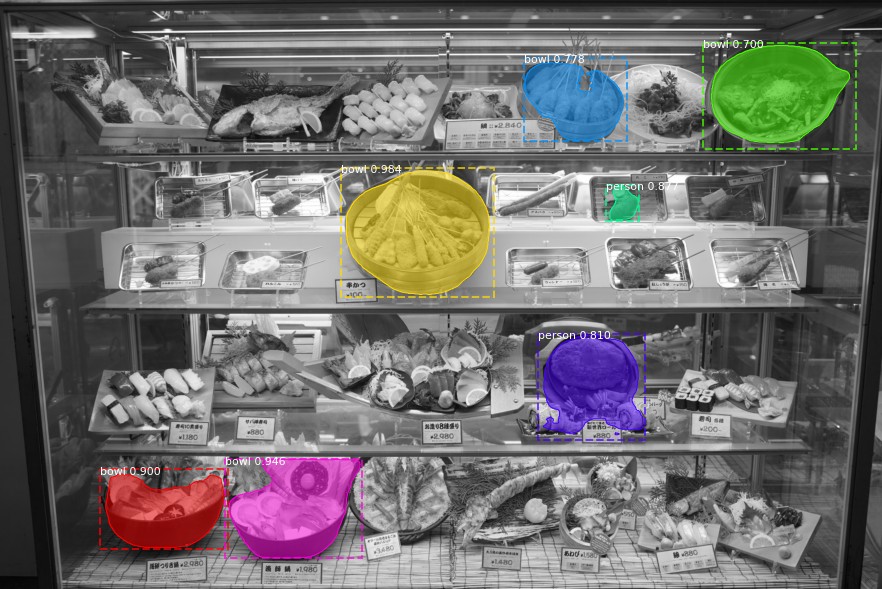}
    \caption{Mask R-CNN trained on COCO. Test image converted to grayscale produces strange detection such as "person" with high confidence.}
    \label{maskRCNNOsakaFood}
\end{figure}

In this section we show several practical and theoretical examples about the consequences of making incorrect but highly confident predictions.

\textbf{What are overconfident predictions?}. A prediction is considered overconfident if it predicts an incorrect class, but the confidence is high. The threshold for an overconfident prediction varies with task and the number of classes. For classification tasks, the maximum probability or confidence is usually used to decide the output class, so any incorrect prediction with confidence higher than $\frac{1}{C}$ where C is the number of classes, can be considered overconfident.

For regression problems, an overconfident prediction is one with low standard deviation or interval length, but the true value is outside of the confidence interval.

\textbf{Examples}. Figure \ref{maskRCNNOkonomiyaki} shows the Mask R-CNN predictions on a test image of an Okonomiyaki, which are incorrectly classified as a Bowl and a Pizza, both with high confidence of 73\% and 78\%. Other objects in this image are also misclassified such as a Spoon and a Fork.

Figure \ref{maskRCNNOsakaFood} also shows Mask R-CNN predictions, but the image was transformed to grayscale, most likely making it out of distribution. Predicted confidences are not a good indicator that this example is far from the training distribution, and even there are some grossly incorrect predictions such as food plates predicted as persons (with confidences 77\% and 81\%).

\textbf{Image Classification}. Misclassified and out of distribution examples cannot be detected by using class confidences \cite{hendrycks2017baseline}.

\textbf{Object Detection}. Modern object detectors produce predictions based on a thresholding class confidence of candidate bounding boxes. Overconfident misclassifications of the background class produce false positives. An important aspect is the variance of bounding boxes which is generally not considered when deciding on final predictions. An overconfident bounding box is one that poorly localizes the object with a low interval length. Most state of the art object detectors do not output any kind of bounding box uncertainty.

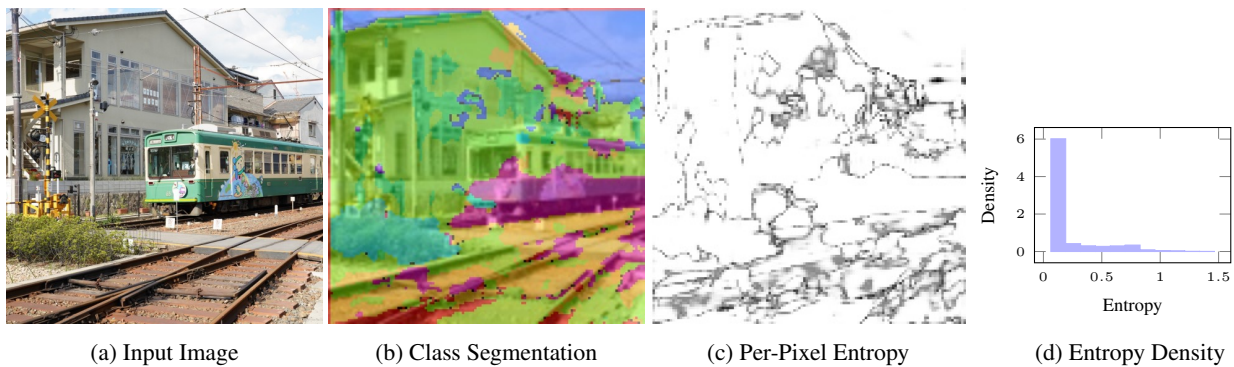
\begin{figure*}[!th]
    \begin{subfigure}[b]{0.24\textwidth}
        \includegraphics[width=\textwidth]{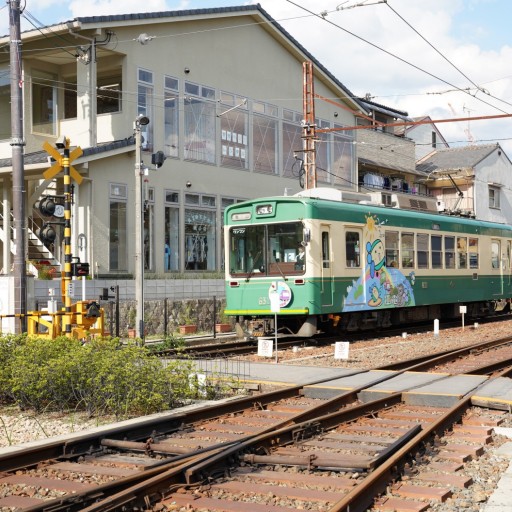}
        \caption{Input Image}
    \end{subfigure}
    \begin{subfigure}[b]{0.24\textwidth}
        \includegraphics[width=\textwidth]{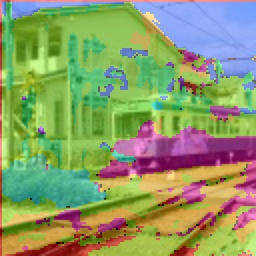}
        \caption{Class Segmentation}
    \end{subfigure}
    \begin{subfigure}[b]{0.24\textwidth}
        \includegraphics[width=\textwidth]{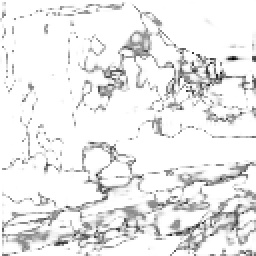}
        \caption{Per-Pixel Entropy}
    \end{subfigure}
    \begin{subfigure}[b]{0.24\textwidth}
        \begin{tikzpicture}
            \begin{axis}[height = 0.15 \textheight, width = \textwidth, xlabel={Entropy}, ylabel={Density}, ymajorgrids=false, xmajorgrids=false, grid style=dashed, legend pos = north east, legend style={font=\scriptsize}, tick label style={font=\tiny}, label style={font=\scriptsize}]
                
                \addplot[mark = none, blue!30,ybar interval,fill] table [x  = bin, y  = density, col sep = semicolon] {images/entropy-density.csv};
                
            \end{axis}
        \end{tikzpicture}
        \caption{Entropy Density}
    \end{subfigure}
    \caption{Semantic segmentation example with U-Net trained on Cityscapes. This railway scene is far from the training set, yet entropy does not indicate that anything is wrong, and as expected, high entropy is present between segmentation region boundaries only, not inside each regions. Entropy values are low for light shades, and high for dark shades.}
    \label{segmentationExample}
\end{figure*}

\textbf{Instance Segmentation}. Binary masks predicted for each instance box by thresholding a set of predicted soft masks, and overconfident predictions can produce inaccurate masks, as seen in Figure \ref{maskRCNNOkonomiyaki}. Most issues are shared with object detectors. An example for the semantic segmentation case is presented in Figure \ref{segmentationExample}.

\textbf{Autonomous Vehicle Perception}. In this task the stakes are higher. Any object that is missed by the object detector can be a potential collision accident, as it has been reported in real-life situations with experimental autonomous vehicles by the US National Transport Safety Board \cite{ntsb2019collision}, due to situations that are not present in the training set. An overconfident prediction could be a pedestrian detection in the incorrect position (sidewalk vs the road), or the production of false detections that could stop the vehicle suddenly or disable autonomous mode.  This shows the importance of out of distribution detection for real-world computer vision applications, as a way to detect situations far from the training set, and provide a signal for the system to be aware and act upon this information. 


\section{The Advantages of Model Uncertainty}

In this section we discuss some advantages that come with properly estimating model uncertainty.

\begin{figure}
    \centering
    
    \begin{tabular}{>{\centering\arraybackslash}p{0.20\linewidth}
                    >{\centering\arraybackslash}p{0.20\linewidth}
                    >{\centering\arraybackslash}p{0.20\linewidth}
                    >{\centering\arraybackslash}p{0.20\linewidth}}
        \textbf{Image} & \textbf{True}  & \textbf{Classic} & \textbf{Model}\\
              & \textbf{Labels} & \textbf{Model}  & \textbf{with}\\
              &                 &                 & \textbf{Uncertainty}\\
        \includegraphics[height=1.5cm]{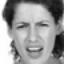} &
        \includegraphics[height=1.5cm]{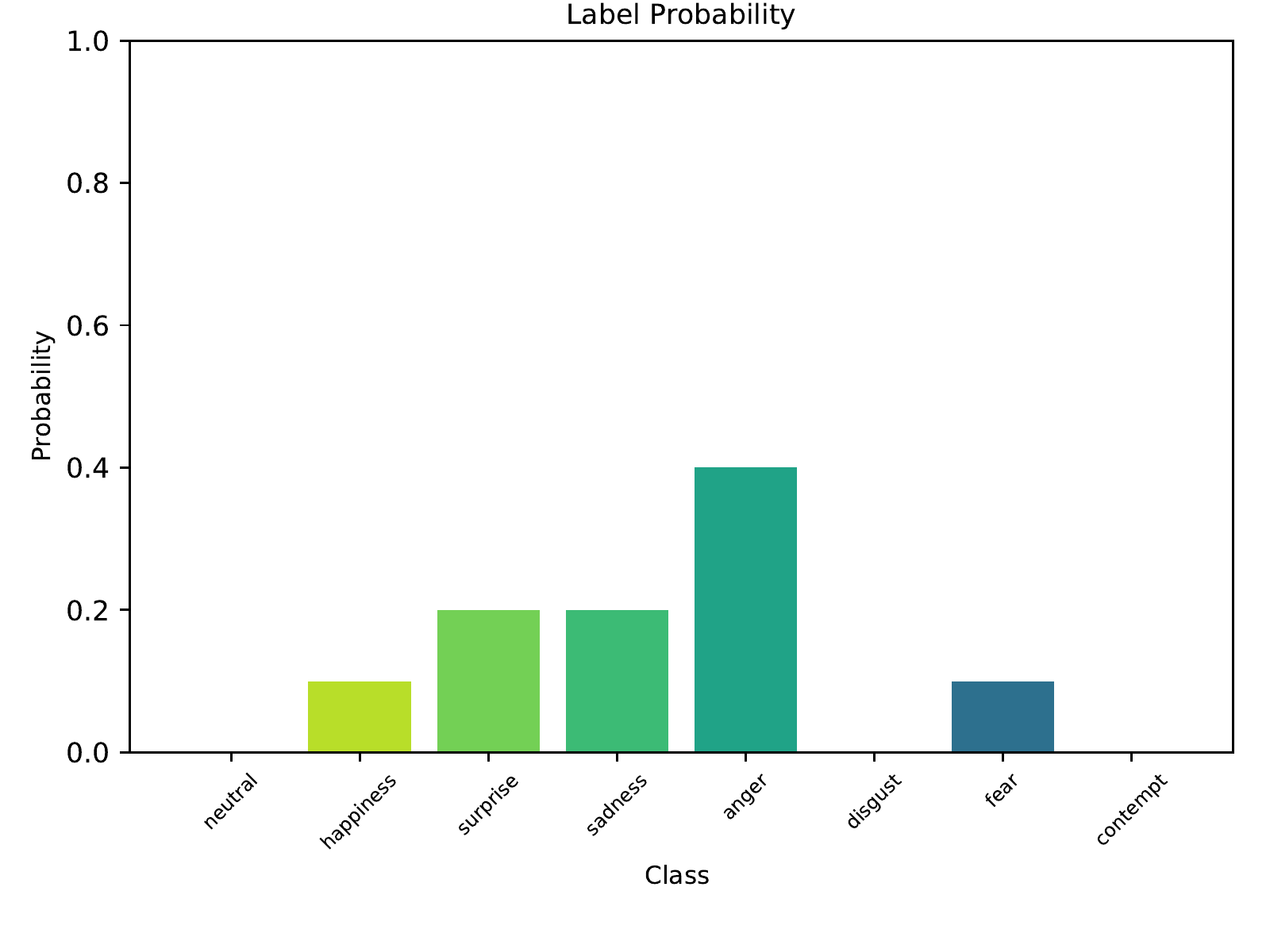} &
        \includegraphics[height=1.5cm]{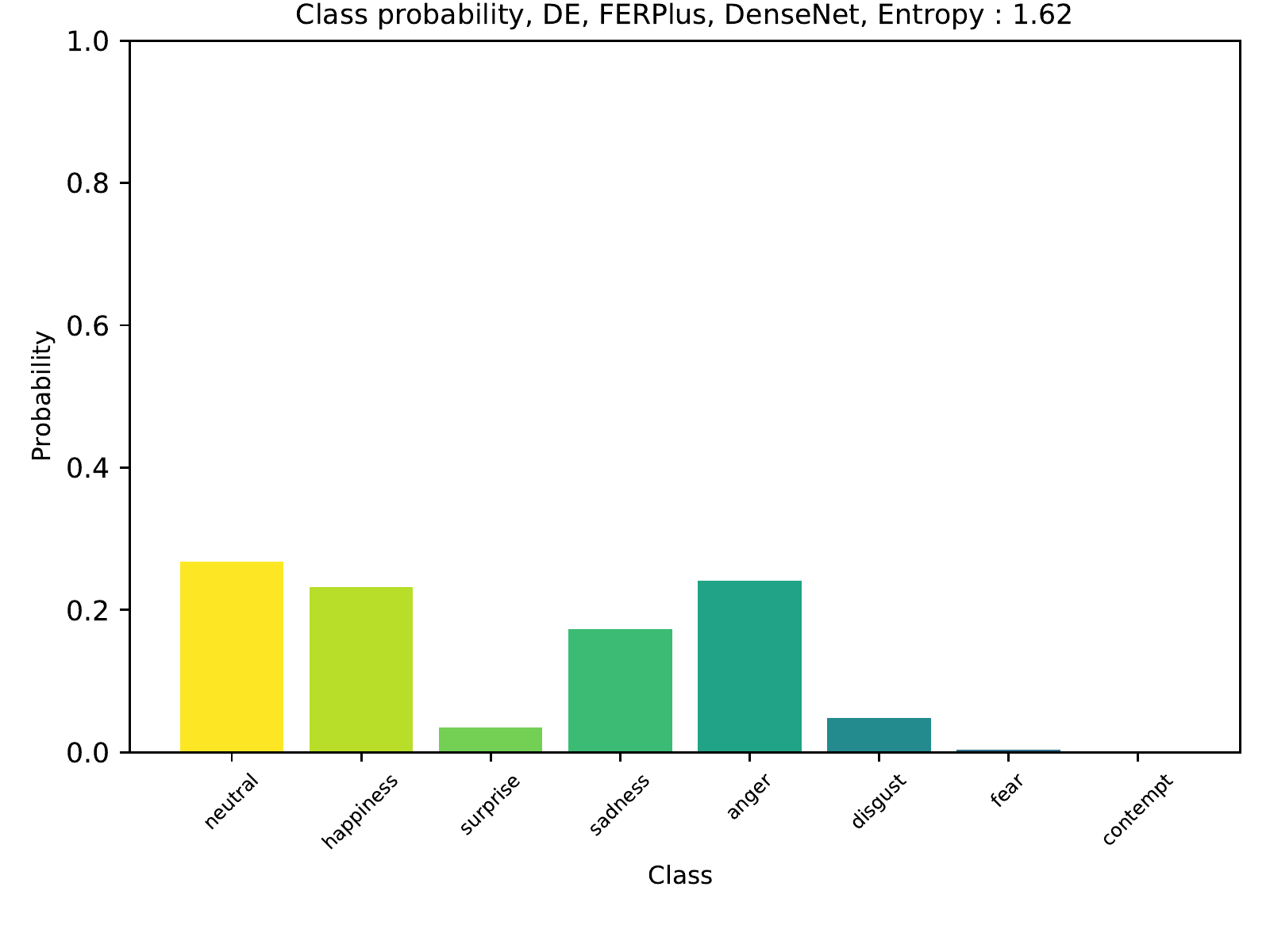} & 
        \includegraphics[height=1.5cm]{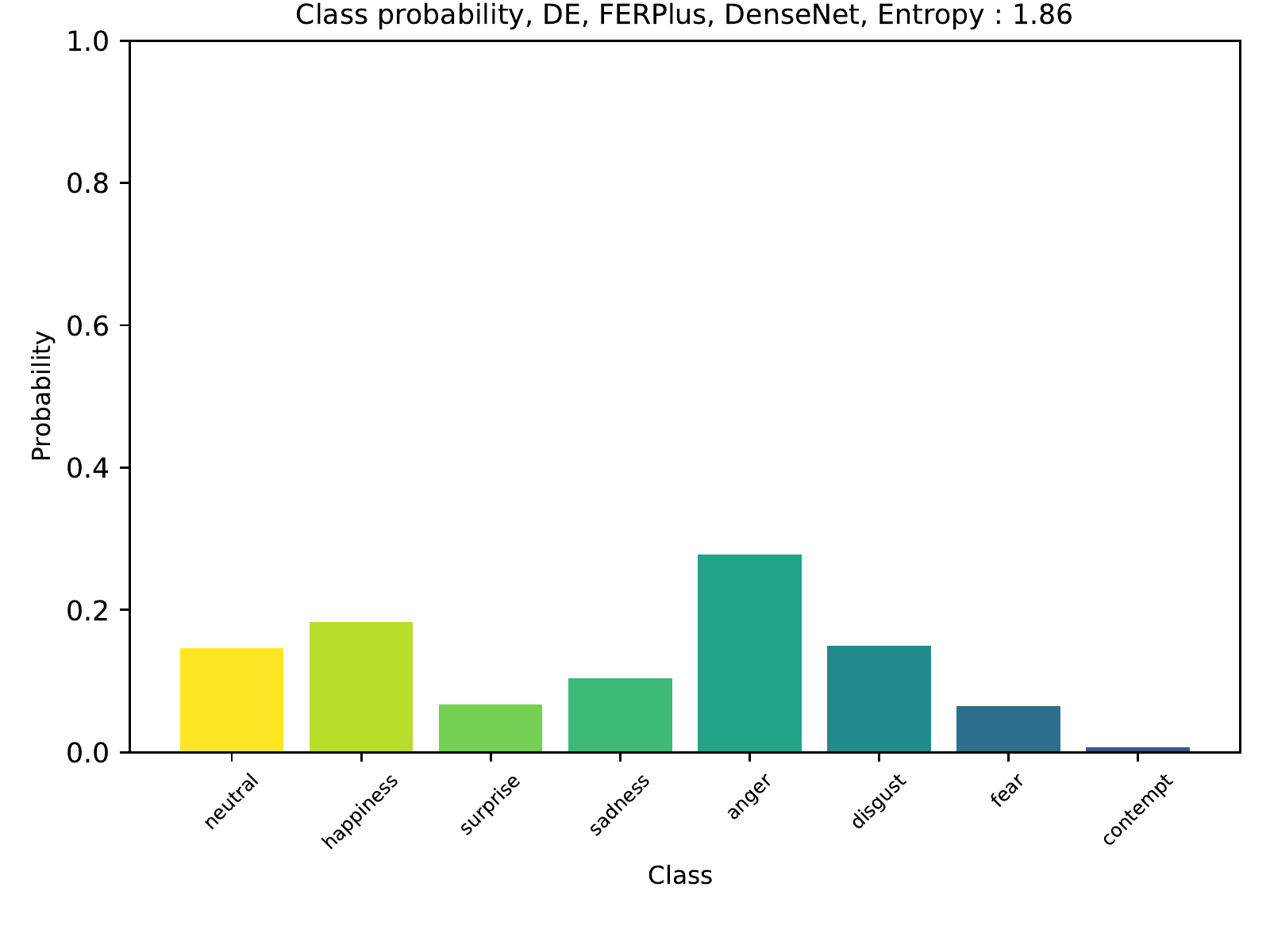}\\
        & Anger & Neutral & Anger\\
        
        \includegraphics[height=1.5cm]{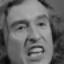} &
        \includegraphics[height=1.5cm]{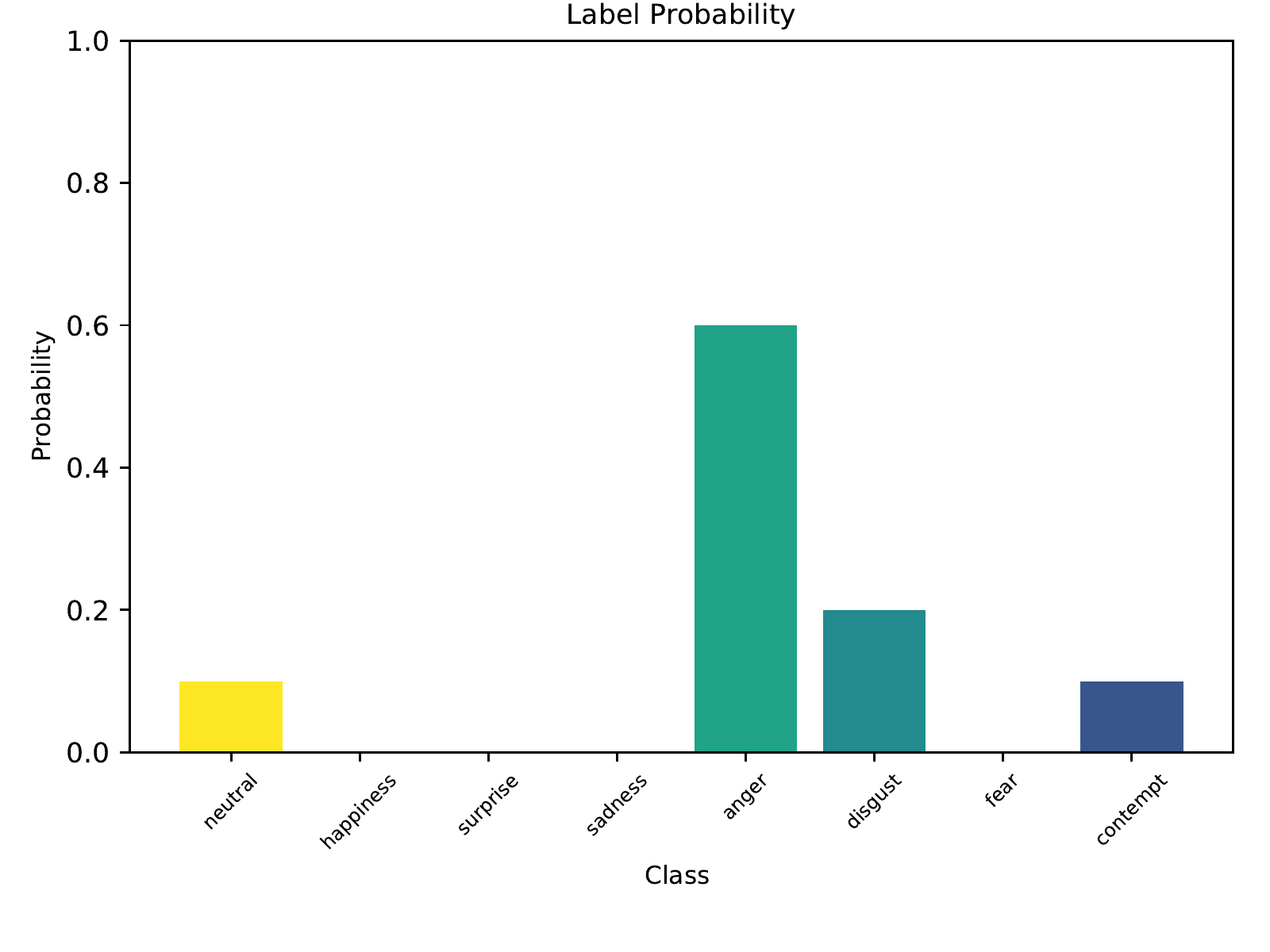} &
        \includegraphics[height=1.5cm]{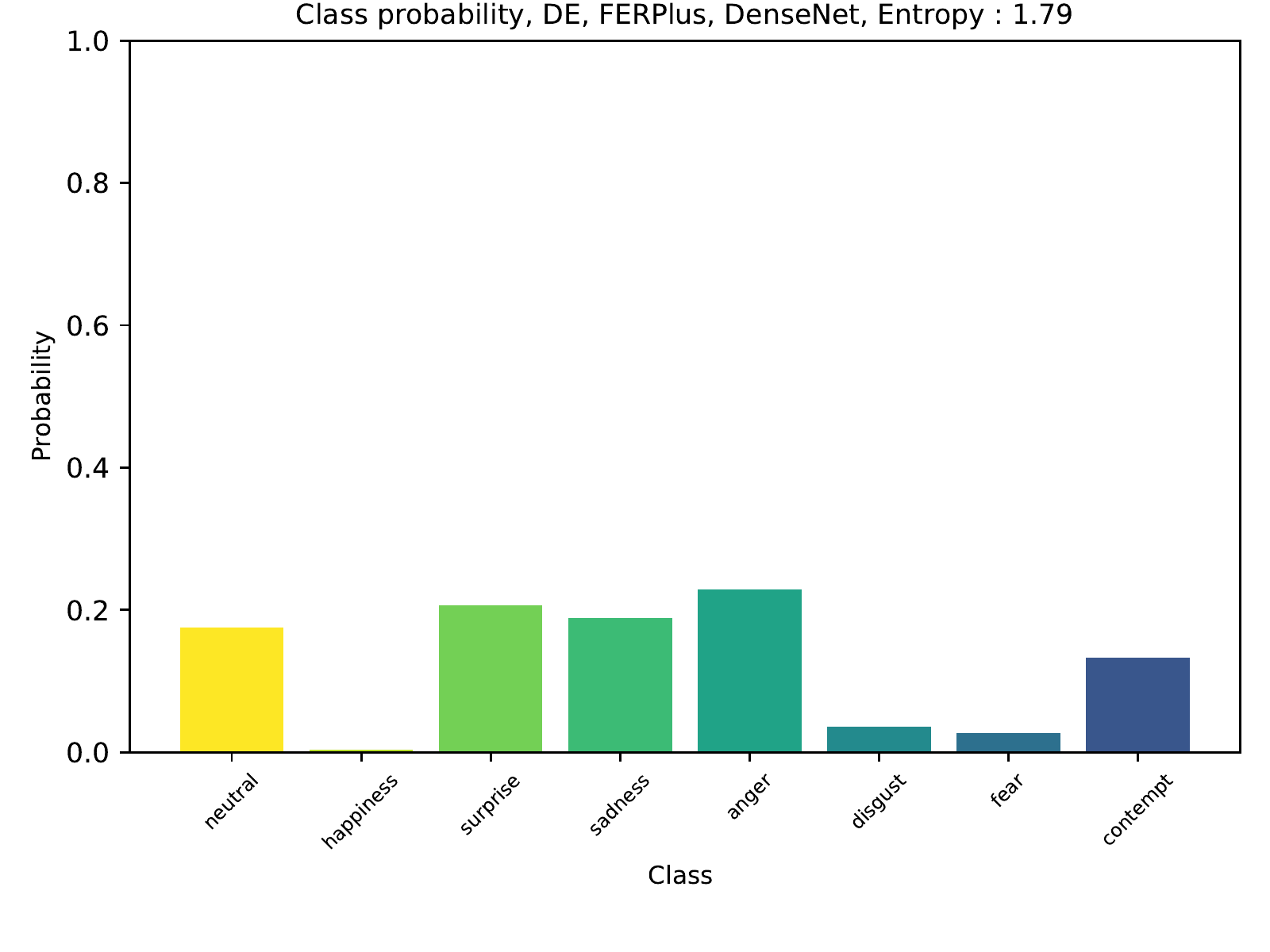} & 
        \includegraphics[height=1.5cm]{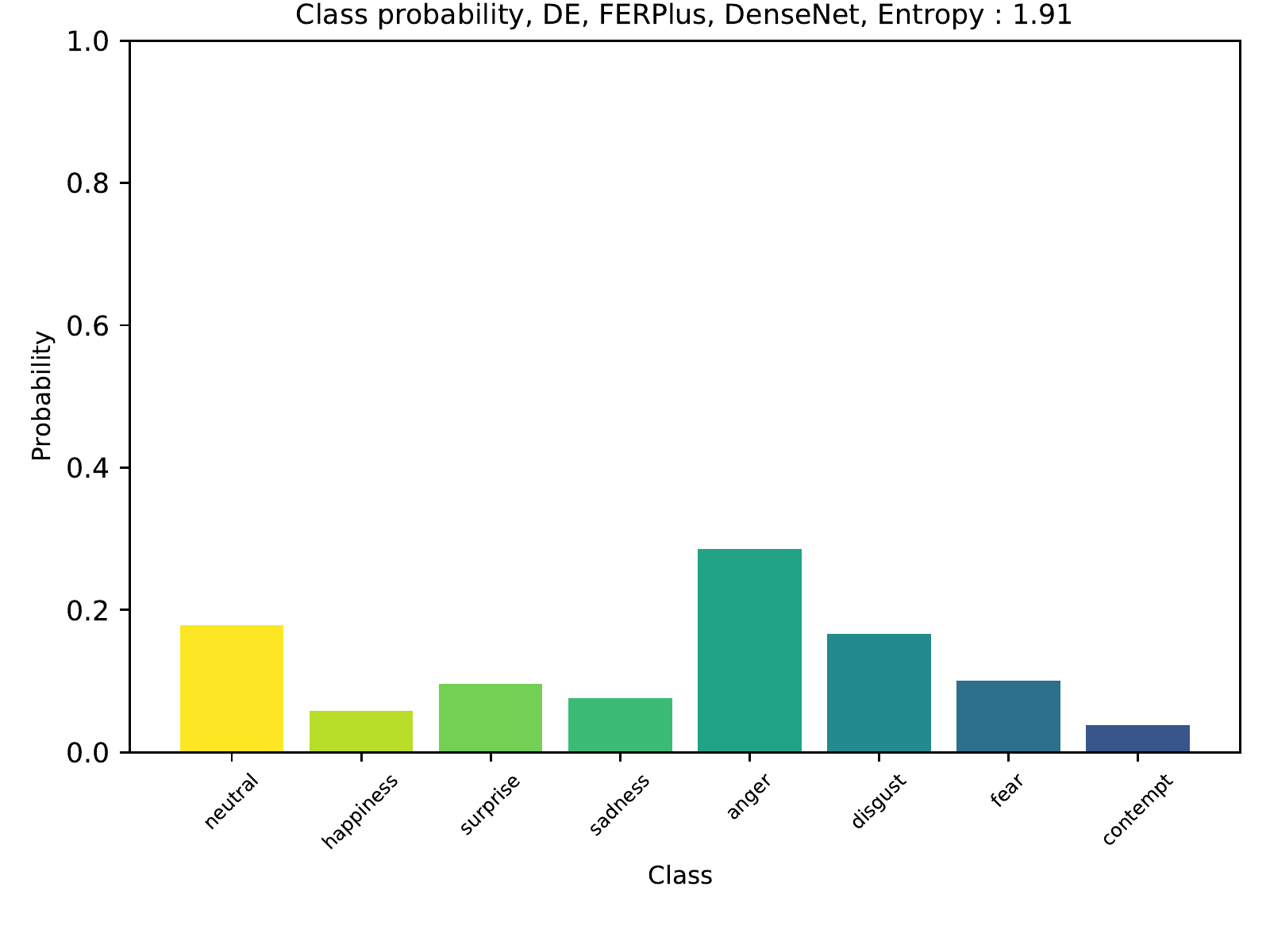}\\
        & Anger & Anger & Anger\\
        
        \includegraphics[height=1.5cm]{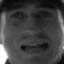} &
        \includegraphics[height=1.5cm]{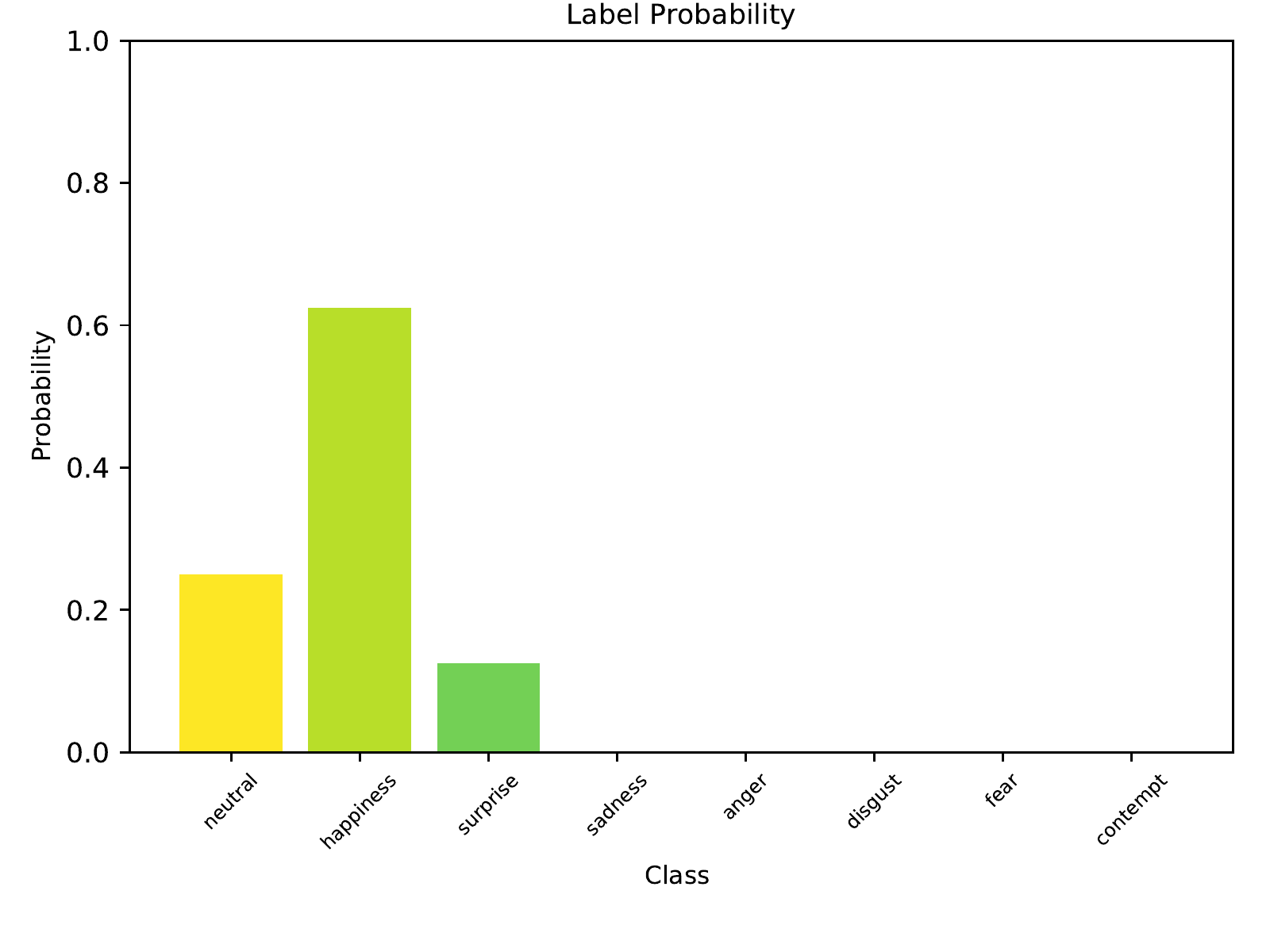} &
        \includegraphics[height=1.5cm]{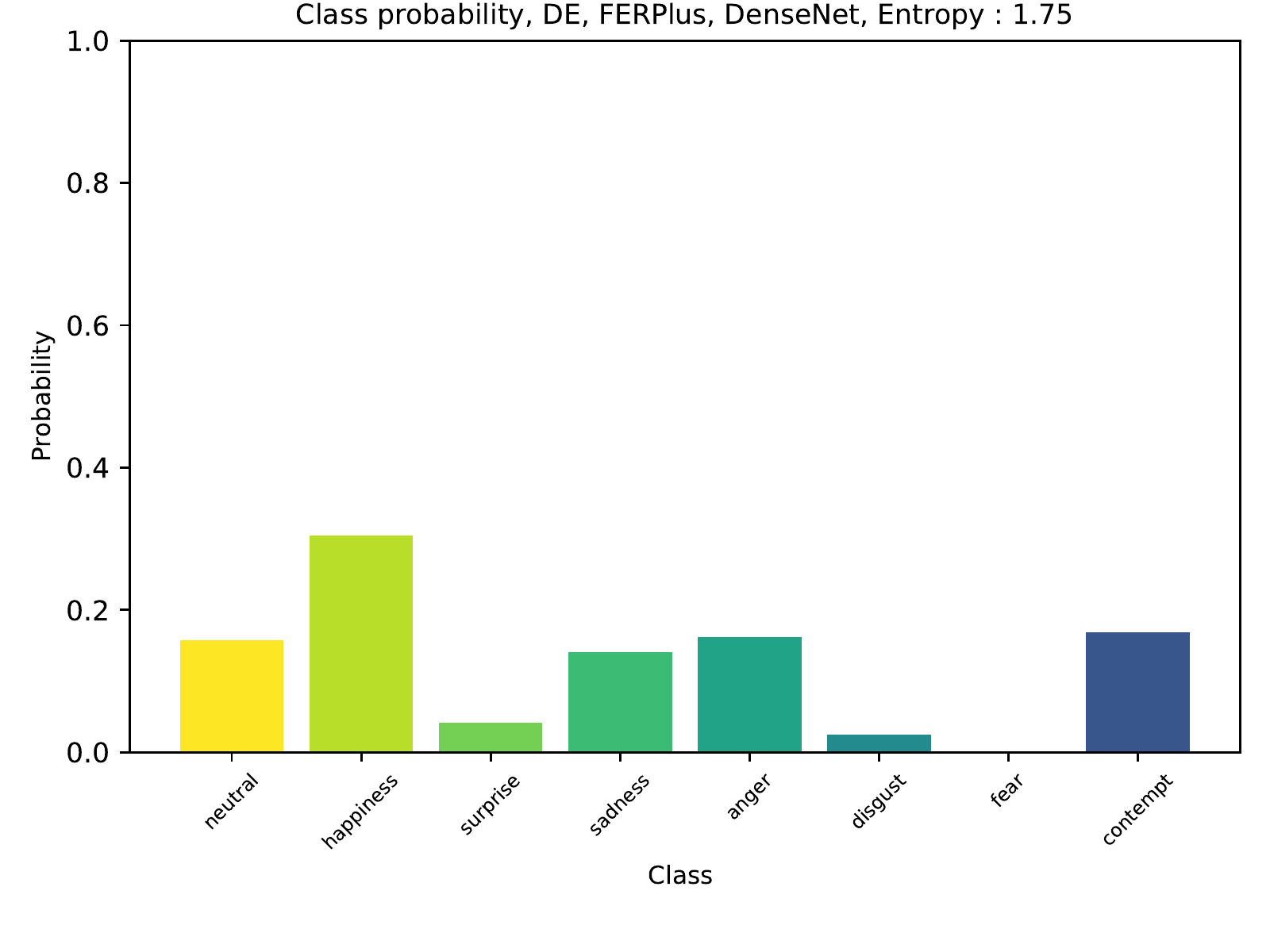} & 
        \includegraphics[height=1.5cm]{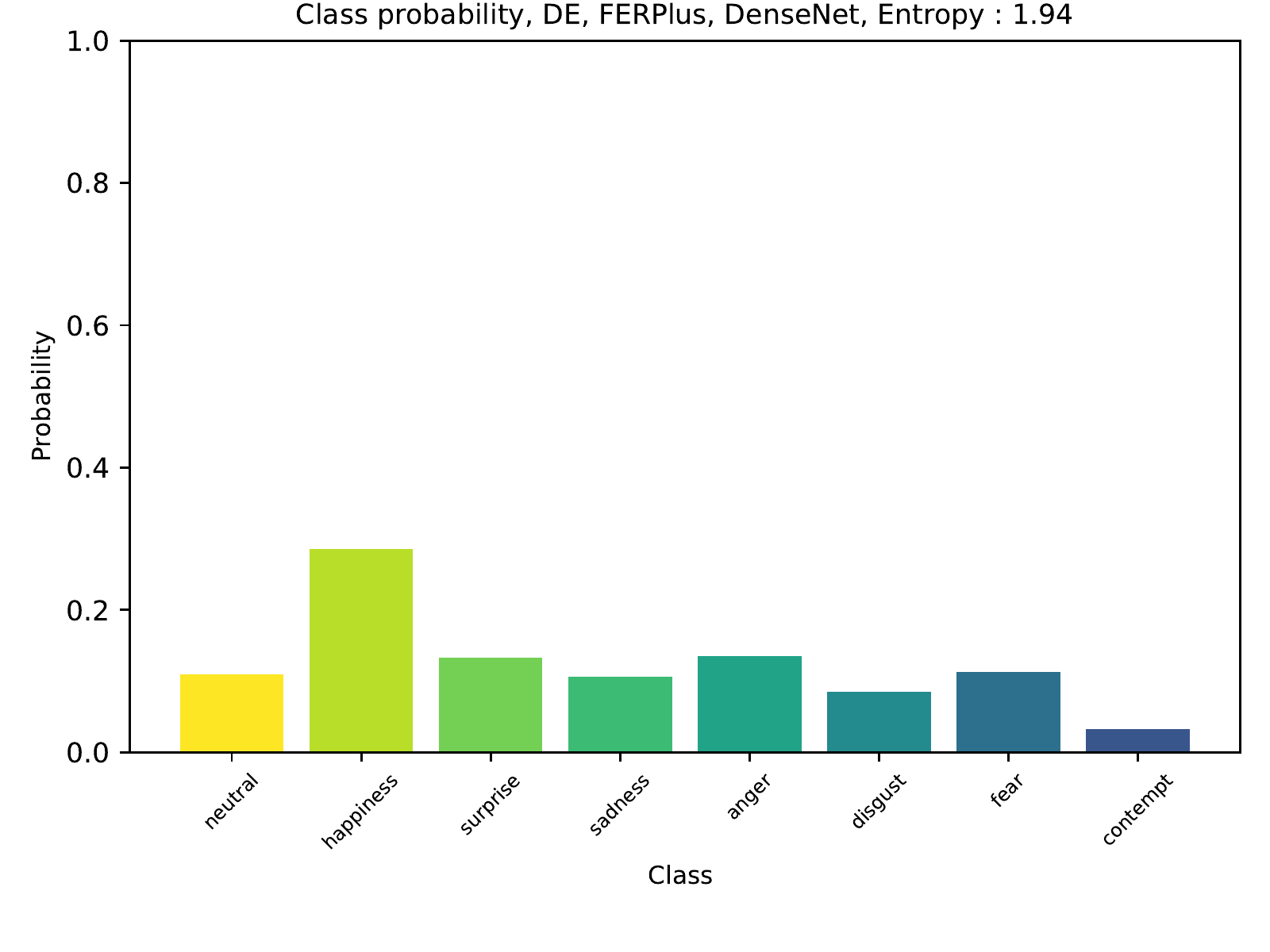}\\
        & Happiness & Happiness & Happiness\\        
        
        \includegraphics[height=1.5cm]{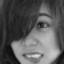} &
        \includegraphics[height=1.5cm]{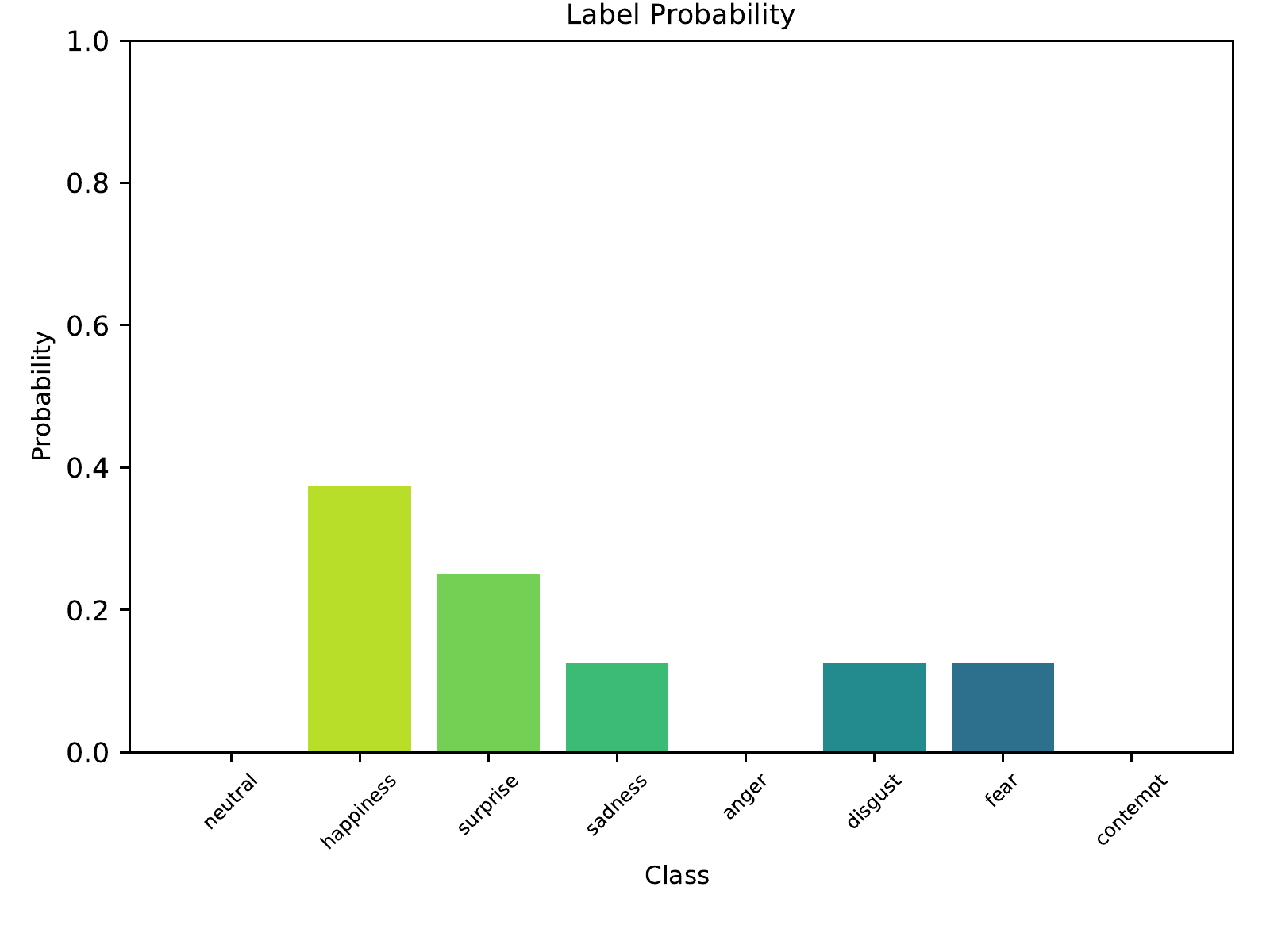} &
        \includegraphics[height=1.5cm]{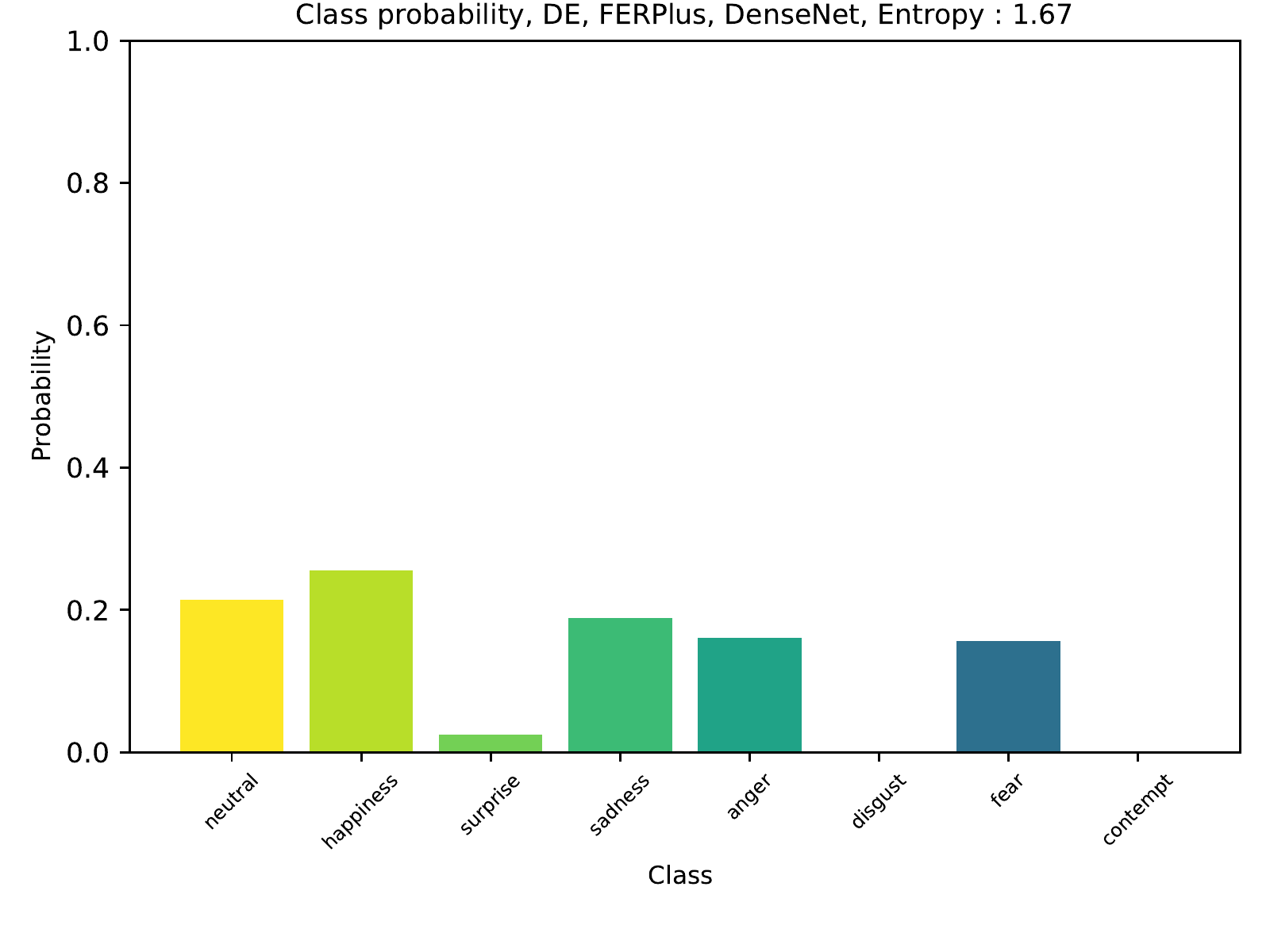} & 
        \includegraphics[height=1.5cm]{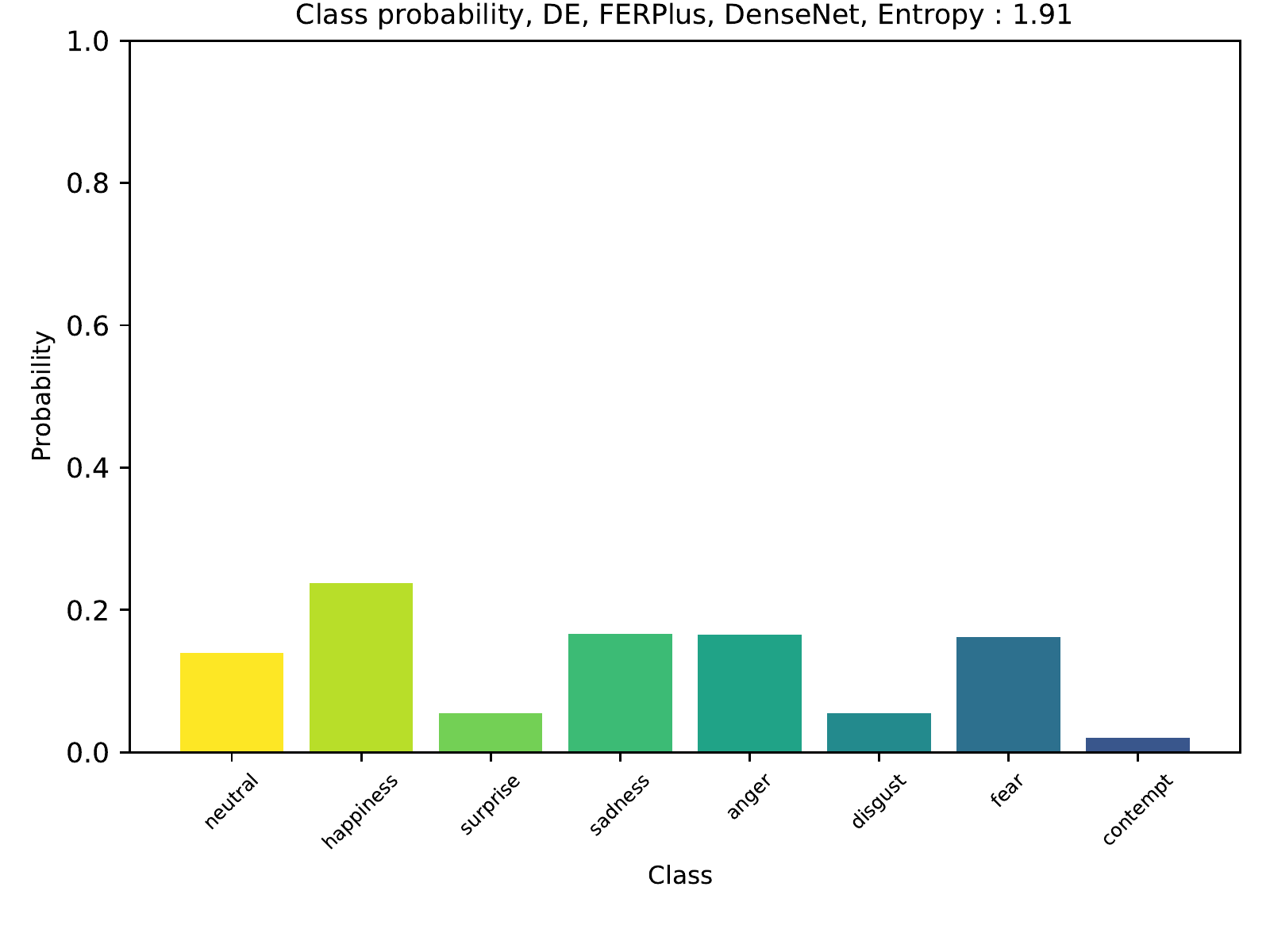}\\
        & Happiness & Happiness & Happiness\\
    \end{tabular}
    \vspace*{0.1em}        
    \caption{Comparison of DenseNet predictions on the FERPlus dataset \cite{barsoum2016training}. This is a high-uncertainty task, as the labels are probability distributions over classes (not one-hot encoded), so there are multiple "correct" classes due to the visual ambiguity in facial expressions. A model with proper epistemic uncertainty produces probabilities that make more sense, are closer to the true probabilities, and are less sparse. Classes are: Neutral, Happiness, Surprise, Sadness, Anger, Disgust, Fear, Contempt. Results adapted from \cite{matin2020}.}
    \label{ferplusComparison}
\end{figure}

\textbf{Tasks with High Uncertainty}. Some tasks have a natural tendency to produce high uncertainty. For example, classes that are visually similar could lead to a model that can easily confuse them, and this effect can be communicated by the model through properly calibrated per-class probabilities. One clear example is emotion recognition \cite{matin2020}, where even humans have difficulties in predicting which emotion is right, and there could be multiple correct emotions being conveyed by a face.  This example is shown in Figure \ref{ferplusComparison}.

\textbf{Out of Distribution Detection}. The final goal of proper epistemic uncertainty quantification is to perform out of distribution detection. In this task model confidence or uncertainty is used to indicate if a particular input sample is far from the distribution of the training set, indicating that the model might be extrapolating, or if this particular example is difficult for the model to make a decision and might be misclassified.

We believe that this is the most important application of uncertainty quantification in machine learning and computer vision models. We should not only put emphasis on model predictions but also in the model confidence, as it allows us to gauge how much a human should trust the prediction. Low confidence predictions will be interpreted by a human as not trustworthy, and development of computer vision models should take this into account.

Out of distribution detection is even more important for applications that involve humans, such as medical decisions, autonomous driving, and robotics.

\section{Practical Issues}

There are a series of reasons why uncertainty quantification is not used in real-world applications.

\textbf{Computational Costs}. We believe that the biggest barrier to the adoption of uncertainty methods in computer vision is their increased computational cost compared to non-uncertainty methods. For example, using MC-Dropout, MC-DropConnect, or ensembles, requires $N$ forward passes of the full model, with the value of $N$ controlling a trade-off between a good posterior distribution approximation, and requiring $N$ times the computation cost of a single model.

\textbf{Approximation Quality}. Most BNNs are implemented using approximations to the true posterior, since computing it is intractable. This introduces additional approximation problems, where the quality of uncertainty that is produced by the model can be poor for some applications \cite{ovadia2019trust}. For the out of distribution detection task, this has motivated to build methods that produce non-uncertainty outputs \cite{schwaiger2020uncertainty} that can still be used for out of distribution detection, like ODIN \cite{hsu2020generalized}.

\section{Future Challenges}

\textbf{A. Computational Cost}. Properly estimating epistemic uncertainty requires the use of weight distributions for Bayesian neural networks, with increased complexity to estimate the posterior predictive distribution through sampling or ensembling. Many methods can estimate aleatoric uncertainty with a single model but this is a trade-off, since epistemic uncertainty is lost. A lot of research is devoted to try to minimize the overhead of uncertainty quantification in modern neural networks.

\textbf{B. Ground Truth Uncertainty}. Most datasets contain labels without uncertainty information on them. Some datasets like PASCAL VOC contain difficulty labels that are used for evaluation purposes. Uncertainty labels would help in the evaluation of probabilistic methods and bias their selection and development into uncertainty that is useful for certain applications.

\begin{figure}[t]
    \includegraphics[width=\linewidth]{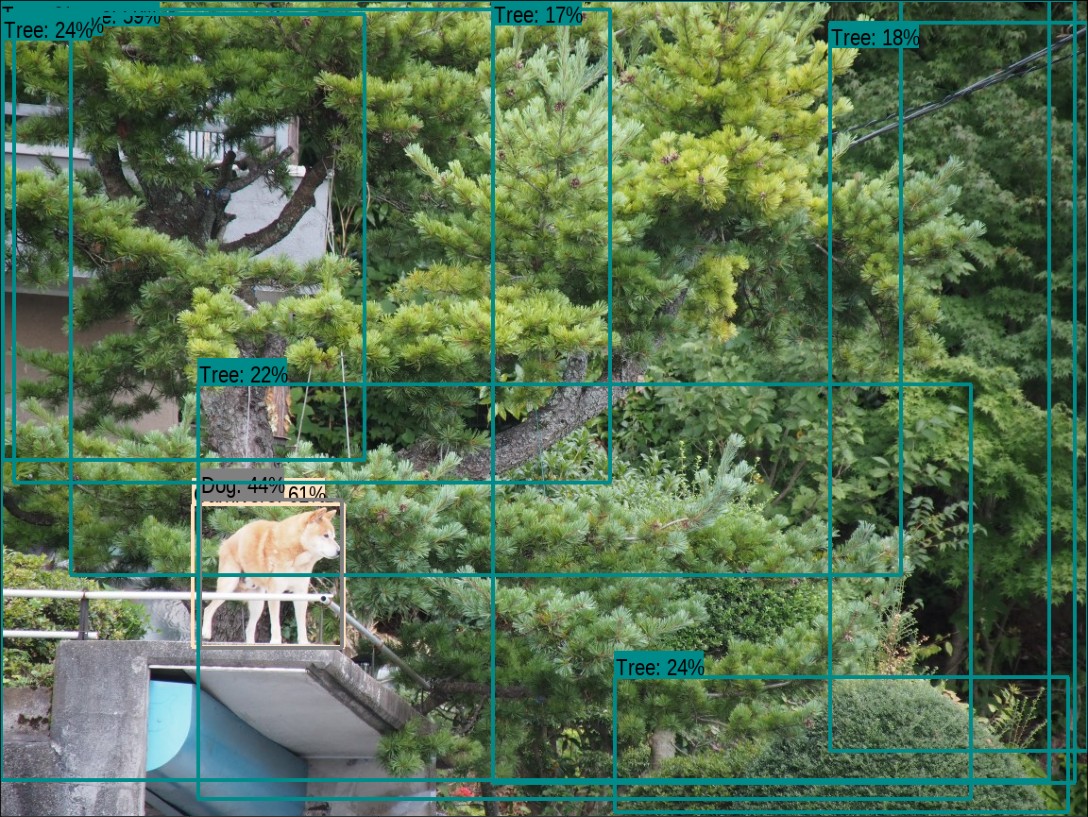}
    \caption{SSD with Inception-ResNet predictions (trained on OpenImages v4). This model seems to think that the dog is 61\% carnivorous and only 44\% dog. This could be due to more images labeled as carnivorous than dog, but the predicted confidences are counter-intuitive from a human perspective.}
    \label{shibaSSD}
\end{figure}

\textbf{C. Human Uncertainty}. Epistemic uncertainty estimated by Bayesian models can be counter-intuitive for humans. An example is shown in Figure \ref{shibaSSD}, where a dog is detected as 44\% dog and 61\% carnivorous, which is a bit strange from the human point of view. Many tree detections with low confidence are also present in the same figure, indicating a problem in localizing the tree in the background.
This motivates the need for models that can produce uncertainty estimates that are closer to what a human would expect. There is already some work in this area like Peterson et al. \cite{peterson2019human}.

\textbf{D. Simulation tools with Uncertainty}. Simulation is a powerful tool to create synthetic data. Building simulation models with uncertainty would allow the automatic production of labels for uncertainty, expanding the usefulness of simulated data. This kind of data would also allow the simulation of dangerous situations (traffic accidents, natural disasters, fires/arson, odd or uncommon scenarios, etc) that cannot be captured in real life, and be used to evaluate out of distribution detection.

\textbf{E. Decision Making with Uncertainty}. A long term challenge is to integrate predictions with uncertainty into the decision making processes of complex systems. A clear example of this is autonomous driving, where an ideal case is to defer driving to a human if the uncertainty in the perception systems is too high, or to use uncertainty in non-perception systems to detect faults and unexpected situations in the road.

\textbf{F. Metrics}. Standard metrics are not aimed at measuring good uncertainty, and there is a need to evaluate uncertainty quality. Metrics like the expected calibration error partially do this job. There should be benchmarks to evaluate the quality of aleatoric and epistemic uncertainty estimation, as well as its disentanglement. There are benchmarks for out of distribution detection.

These metrics also need to be considered during model development, and particularly when reporting a new model in the literature.

\textbf{G. Out of Distribution and Domain Shift Performance}. Last but not least, methods that produce high quality epistemic uncertainty estimates can still fail in certain out of distribution and domain shift scenarios. Ovadia et al. has shown how in a simple image classification setting, uncertainty quality degrades with synthetic corruptions of test images \cite{ovadia2019trust}. Additional research \cite{schwaiger2020uncertainty} points in the direction that the quality of uncertainty should be a target for improvement.

\textbf{H. Openness}. One important component of trustworthy AI is that models and methods that are used in practice, can be accountable, explainable, and overall they can be trusted. But none of this is possible if we do not know which models and methods are being used by applications, and important details such as training configuration, dataset characteristics, or neural network architectures. These details are important to quantify the quality of uncertainty that these methods produce, and their out of distribution detection capabilities. Regulation will probably require such information for certification of AI systems that are used in public spaces (such as autonomous vehicles).

\section{Conclusions and Future Work}

This paper presented a meta-review of real world applications using computer vision, and note that most of the models that are used in the state of the art, do not have proper uncertainty quantification. We discussed common issues that arise when selecting and using models with uncertainty, and bad practices made in the literature.

We expect that out work motivates the community to move away from classical neural networks into using Bayesian neural networks and models with proper epistemic uncertainty for real-world applications, and to evaluate performance in out of distribution detection tasks, in order to ensure the safe use of machine learning models.

In the future we should see more computer vision applications that can tell the user when they do not know the answer or predict that are operating outside of their training capabilities, and we believe that this will go in the direction of safe and trustworthy artificial intelligence.

\section*{Acknowledgments}

We would like to thank Octavio Arriaga for access to a U-Net segmentation model, and the anonymous reviewers for their insightful comments.

{\small
\bibliographystyle{ieee_fullname}
\bibliography{biblio}
}

\end{document}